%% file: anonymous-submission-latex-2026.tex
\documentclass[letterpaper]{article} 
\usepackage{aaai2026}  
\usepackage{times}  
\usepackage{helvet}  
\usepackage{courier}  
\usepackage{hyperref}
\usepackage{graphicx} 
\usepackage{natbib}  
\usepackage{caption} 
\usepackage{algorithm}
\usepackage{algorithmic}

\usepackage{graphicx}
\usepackage{booktabs}
\usepackage{algorithm}  
\usepackage{algorithmic}
\usepackage{multicol}
\usepackage{multirow}
\usepackage[normalem]{ulem}
\useunder{\uline}{\ul}{}
\usepackage{amsmath}
\usepackage{cleveref}
\usepackage{subcaption}
\usepackage{tabularx}   
\usepackage{amsfonts}       

\newcommand{\mcX}{\mathcal{X}}
\newcommand{\mcY}{\mathcal{Y}}
\newcommand{\mcZ}{\mathcal{Z}}
\newcommand{\mcD}{\mathcal{D}}
\newcommand{\mcN}{\mathcal{N}}
\newcommand{\boldx}{\boldsymbol{x}}
\newcommand{\boldy}{\boldsymbol{y}}
\newcommand{\boldz}{\boldsymbol{z}}
\newcommand{\boldc}{\boldsymbol{c}}
\newcommand{\boldmu}{\boldsymbol{\mu}}
\newcommand{\boldsigma}{\boldsymbol{\sigma}}
\newcommand{\boldeps}{\boldsymbol{\epsilon}}
\newcommand{\boldtheta}{\boldsymbol{\theta}}
\newcommand{\boldvarphi}{\boldsymbol{\varphi}}
\newcommand{\boldI}{\boldsymbol{\rm I}}
\newcommand{\balpha}{\bar{\alpha}}
\newcommand{\OURS}{RiskyDiff}

\usepackage{newfloat}
\usepackage{listings}
\DeclareCaptionStyle{ruled}{labelfont=normalfont,labelsep=colon,strut=off} 
\floatstyle{ruled}
\newfloat{listing}{tb}{lst}{}
\floatname{listing}{Listing}
\title{Generating Risky Samples with Conformity Constraints via Diffusion Models}
\author{
    Han Yu\textsuperscript{\rm 1}, 
    Hao Zou\textsuperscript{\rm 1},
    Xingxuan Zhang\textsuperscript{\rm 1},
    Zhengyi Wang\textsuperscript{\rm 1},
    Yue He\textsuperscript{\rm 2}, 
    Kehan Li\textsuperscript{\rm 1}, 
    Peng Cui\textsuperscript{\rm 1}\thanks{Corresponding author}
}
\affiliations{
    \textsuperscript{\rm 1}Tsinghua University, \textsuperscript{\rm 2}Renmin University of China

    yuh21@mails.tsinghua.edu.cn, ahio@163.com, xingxuanzhang@hotmail.com, wang-zy21@mails.tsinghua.edu.cn\\
    hy865865@gmail.com, lkh20@mails.tsinghua.edu.cn, cuip@tsinghua.edu.cn
 
}

\usepackage{bibentry}

\begin{document}

\maketitle

\input{section/0_abstract}

\input{section/1_intro}

\input{section/2_related}

\input{section/3_1_pre}

\input{section/3_2_method}

\input{section/4_exp}

\input{section/5_con}

\section*{Acknowledgements}

This work was supported by Tsinghua-Toyota Joint Research Fund, NSFC (No. 62425206, 62141607), and Beijing Municipal Science and Technology Project (No. Z241100004224009).
Peng Cui is the corresponding author. All opinions in this paper are those of the authors and do not necessarily reflect the views of the funding agencies.

\onecolumn

\appendix

\input{section/X_suppl}

\bibliography{aaai2026}

\end{document}

%% file: section/0_abstract.tex
\begin{abstract}

Although neural networks achieve promising performance in many tasks, they may still fail when encountering some examples and bring about risks to applications. To discover risky samples, previous literature attempts to search for patterns of risky samples within existing datasets or inject perturbation into them. Yet in this way the diversity of risky samples is limited by the coverage of existing datasets.
To overcome this limitation, recent works adopt diffusion models to produce new risky samples beyond the coverage of existing datasets. However, these methods struggle in the conformity between generated samples and expected categories, which could introduce label noise and severely limit their effectiveness in applications. 
To address this issue, we propose \textit{\OURS}~that incorporates the embeddings of both texts and images as implicit constraints of category conformity. 
We also design a conformity score to further explicitly strengthen the category conformity, as well as introduce the mechanisms of embedding screening and risky gradient guidance to boost the risk of generated samples. 
Extensive experiments reveal that \OURS~greatly outperforms existing methods in terms of the degree of risk, generation quality, and conformity with conditioned categories. We also empirically show the generalization ability of the models can be enhanced by augmenting training data with generated samples of high conformity.
\end{abstract}

%% file: section/1_intro.tex
\section{Introduction}
\label{sec:intro}
Deep learning have exhibited impressive power and achieved promising performance~\cite{dosovitskiy2020image, tian2025yolov12} across various applications, where computer vision is a representative. However, models are still susceptible to some risky samples when encountering high-stake applications, such as health care~\cite{irvin2019chexpert,feuerriegel2024causal} and autonomous driving~\cite{lang2019pointpillars,li2022coda}. Thus discovering these risky samples is crucial to unveil the vulnerability of models and enhance their performance by augmenting the datasets.

Previously, a series of works named error slice discovery~\cite{d2022spotlight, eyuboglu2022domino} propose to find coherent patterns of risky samples from an existing dataset. 
Besides, adversarial attack~\cite{DBLP:journals/corr/SzegedyZSBEGF13, fletcher2000practical, DBLP:journals/corr/GoodfellowSS14,madry2017towards} aims to inject perturbations into existing data to fool the models. 
Nevertheless, these solutions are only capable of finding risky samples within or close to the coverage of existing data, which severely limits the diversity of risky samples. 
It is of urgent need to directly generate risky samples beyond the finite empirical data.

Recent development of diffusion models~\cite{ho2020denoising, song2021denoising,dhariwal2021diffusion,ho2022classifier} presents an opportunity to achieve this target via image generation. 
Some works~\cite{chen2023advdiffuser,dai2024advdiff} try to take advantage of diffusion models to directly generate risky samples beyond the coverage of existing reference datasets, but the ground-truth category labels are not guaranteed to keep unchanged during the generation process.
In the standard conditional generation process via advanced diffusion models, the conformity between the generated samples and their expected category labels could mostly be guaranteed for common objects. 
\begin{figure}[tb]
\vspace{-15pt}
  \centering
  \includegraphics[width=\linewidth]{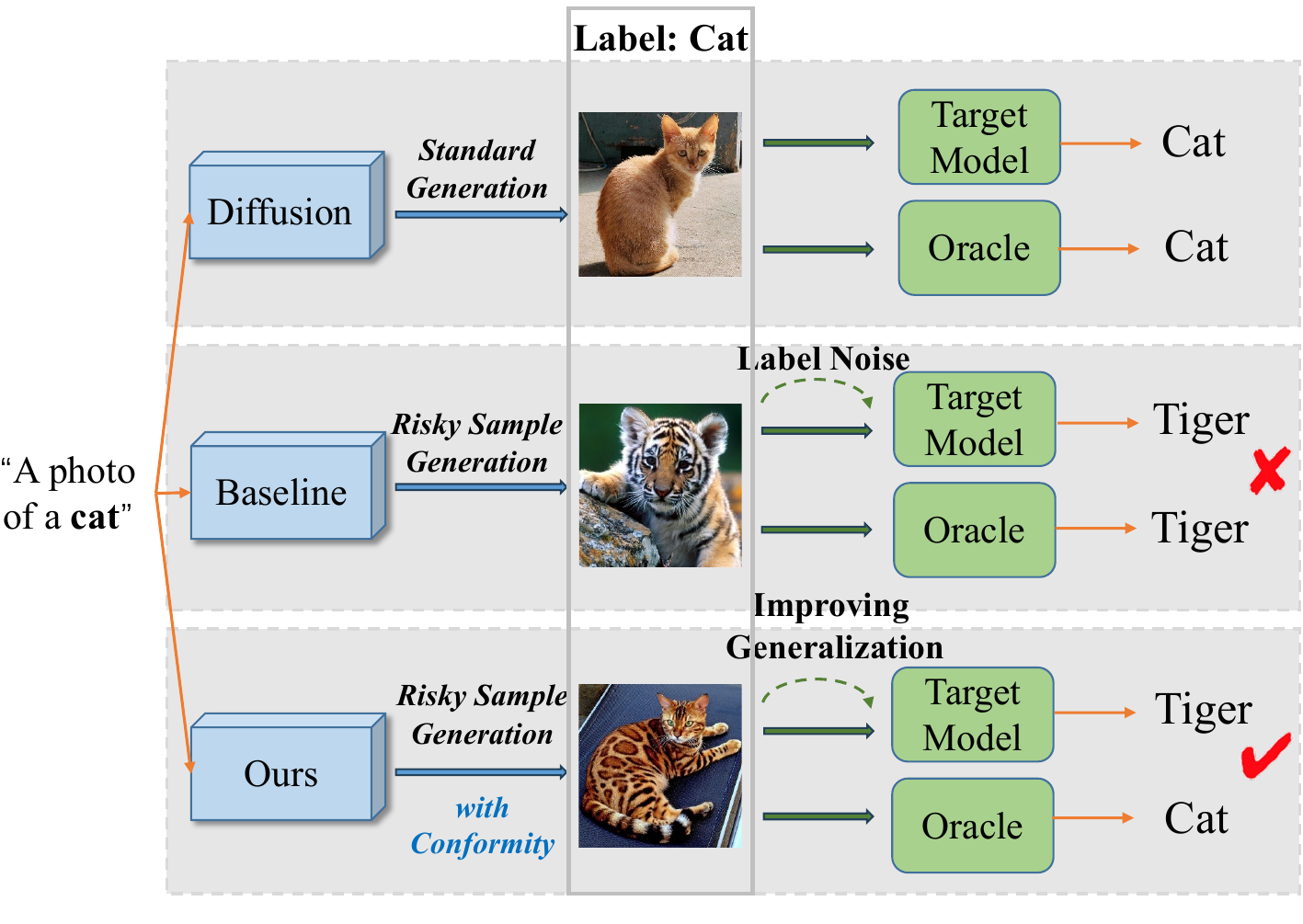}
\vspace{-10pt}
\captionsetup{font=small} 
  \caption{Illustration of generating risky samples with conformity constraints. We require generated samples to deceive the target model but conform to the conditioned category, so that generated samples could improve generalization of the target model after being added to training data. Otherwise, there could be severe label noise in generated samples.}
  \vspace{-15pt}
  \label{fig:paradigm}
\end{figure}
However, when attempting to generate risky samples, the image generation process is guided towards difficult or rare cases, where diffusion models might not be guaranteed to keep the expected category label unchanged. 
This could lead to failure of making generated samples conform to the conditioned category. 
Such nonconformity introduces label noise into generated samples, which could severely limit their application in customized generation demand and hinder their usefulness in augmenting original datasets to improve model performance. 

To mitigate the above issues, we leverage diffusion models as the generative backbone and propose \textbf{\OURS} that utilizes both image and text embeddings as implicit constraints of category conformity. 
We also introduce guidance of an extra conformity score that serves as an explicit constraint of conformity. 
Meanwhile, we introduce mechanisms of embedding screening and risky gradient guidance to increase the risk of generated samples, i.e. the proportion of generated samples that the classification model makes mistakes on. 
Extensive experiments on various image classification datasets validate the effectiveness of our method. The results confirm that \OURS~outperforms existing baselines in terms of the degree of risk, generation quality, and conformity to the conditioned categories. We also demonstrate that the generalization performance of target models can be improved by augmenting original training datasets with generated risky samples of high conformity.
Our contributions are summarized below:
\begin{itemize}
    \item We incorporate both text and image embeddings as input conditions, as well as a conformity score, to encourage category conformity when generating risky samples via diffusion models. 
    \item We introduce the mechanisms of embedding screening and risky gradient guidance to promote the risk of the generated samples with respect to target models. 
    \item Experiments show that our method contributes to higher conformity to expected categories, along with higher risk and generation quality. By adding generated high conformity samples to the training datasets, the target models accomplish improved performance of generalization.
\end{itemize}

%% file: section/2_related.tex
\section{Related Work}
\label{sec:related}

\paragraph{Error Slice Discovery}
Error slice discovery aims to find the interpretable subset of validation data where a model underperforms. They emphasize the coherence of the discovered risky samples. 
Spotlight~\citep{d2022spotlight} aims to learn a sphere as the risky region. 
Domino~\citep{eyuboglu2022domino} adapts the Gaussian mixture algorithm to an error-aware version.
InfEmbed~\citep{wang2023error} uses influence functions as representations of test samples before clustering. 
PlaneSpot~\citep{plumb2023towards} adopts a combination of the dimension-reduced representation and the prediction probability before clustering.
Recently~\citet{yu2025error} leverage the data geometry property to formulate the optimization objective.

\paragraph{Adversarial attack}
Adversarial attack~\citep{DBLP:journals/corr/SzegedyZSBEGF13,DBLP:journals/corr/GoodfellowSS14,miyato2018virtual,kurakin2016adversarial} adds a perturbation with a small norm to the image and deceive deep learning models. 
When model parameters and gradients are not available, black-box attacks~\cite{dong2021query,cheng2024efficient,ilyas2019prior} are investigated.
To generate adversarial examples beyond the $\ell_p$ norm constraints, unrestricted adversarial attacks typically operate on the semantic content of the image, such as color, shape, and texture~\cite{alaifariadef2019, bhattad2020unrestricted, shamsabadi2020colorfool}.
However, overdependence on reference images limits the coverage of potential adversarial examples.

\paragraph{Diffusion models}
\citet{ho2020denoising} firstly demonstrate the impressive capability of diffusion models~\cite{sohl2015deep} in generating images of high quality. It reverses a diffusion process to recover images by denoising noisy data. To accelerate the generation process, denoising diffusion implicit models (DDIM)~\cite{song2021denoising} is proposed to reduce the required inference steps and improve sampling efficiency. \citet{dhariwal2021diffusion} modify the model architecture to improve generation quality and introduce gradients of extra classifiers into the sampling process to achieve conditional generation. 
To eliminate the requirement on extra classifiers, classifier-free guidance~\cite{ho2022classifier} is proposed. To strengthen the flexibility in controlling the content, Stable Diffusion~\cite{rombach2022high} and Stable-unCLIP~\cite{ramesh2022hierarchical} are trained to produce images with user-specified
text prompts.

%% file: section/3_1_pre.tex
\section{Method}

\subsection{Preliminaries}
\label{sec:pre}

\begin{figure*}[tb]
  \centering
  \includegraphics[width=\linewidth]{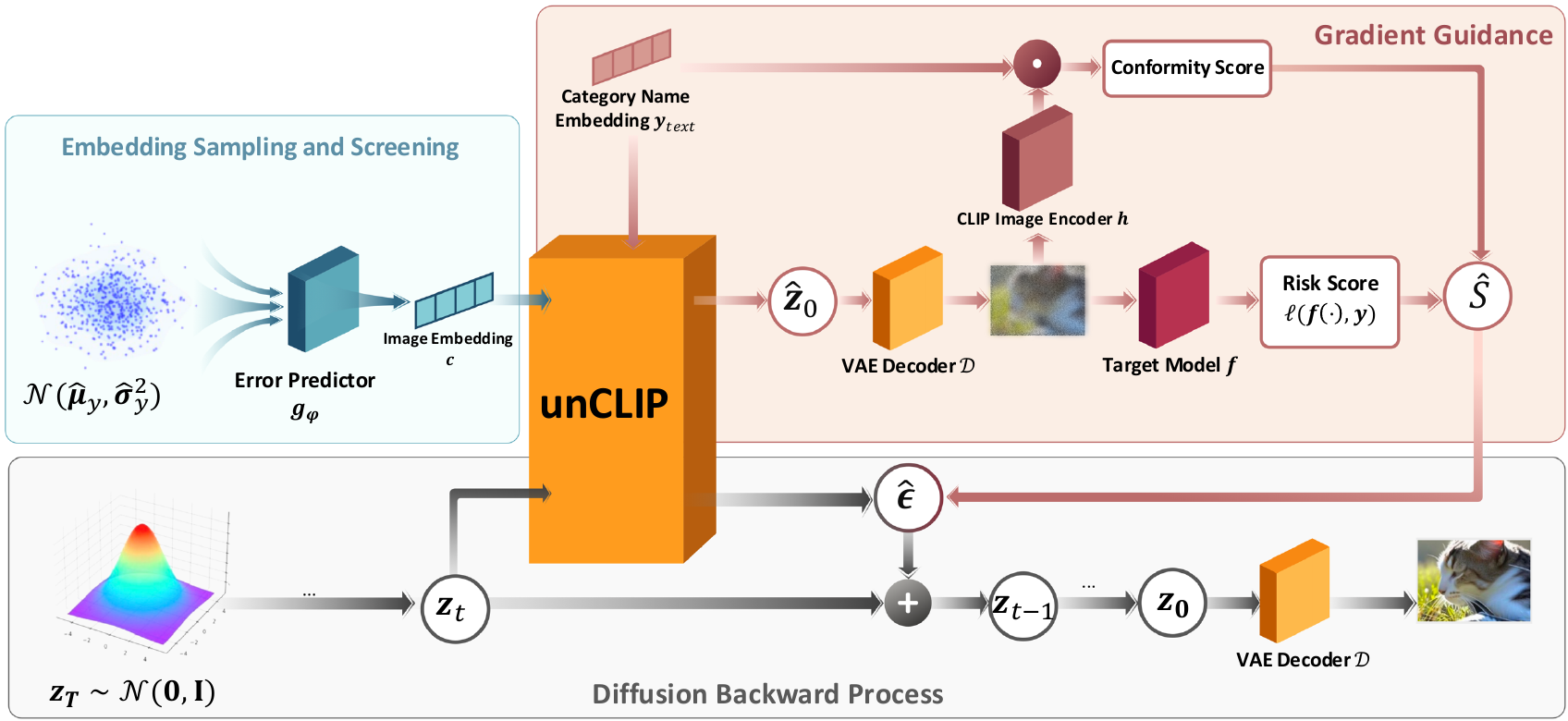}
  \caption{Overall framework of \OURS. The majority of this figure shows a single backward step of the sampling process.}
  \vspace{-10pt}
  \label{fig:pipeline}
\end{figure*}

Throughout the paper we use $[N]$ to represent $\{1,2,...,N\}$.
Let $f: \mcX\mapsto\mcY$ be an image classification model, where $\mcY$ is the category label space.

\paragraph{Diffusion models} The forward process of diffusion is a Markov chain with the transition probability function as $q(\boldx_t|\boldx_{t-1})=\mcN(\boldx_t; \sqrt{\alpha_t}\boldx_{t-1}, (1-\alpha_t)\boldI)$. A nice property of it is $q(\boldx_t|\boldx_0)=\mcN(\boldx_t; \sqrt{\balpha_t}\boldx_0, (1-\balpha_t)\boldI)$, where $\balpha_t=\Pi_{i=1}^t\alpha_i$.  
The objective for a diffusion generative model is to simulate the backward diffusion process, i.e. sampling $\boldx_{t-1}$ based on $\boldx_{t}$. Thus a noise predictor is trained by minimizing the loss $\mathbb{E}_{\boldx_0, \boldeps\sim\mcN(0,\boldI), t\in [T]}||\boldeps-\boldeps_{\boldsymbol{\theta}}(\sqrt{\balpha_t}\boldx_0+\sqrt{1-\balpha_t}\boldeps, t)||^2$~\citep{ho2020denoising}. 
Given the trained noise predictor, the sampling process of DDIM~\citep{song2021denoising} is 
\begin{equation}
\small
    \boldx_{t-1}=\sqrt{\balpha_{t-1}}(\frac{\boldx_t-\sqrt{1-\balpha_t}\boldeps_{\boldsymbol{\theta}}(\boldx_t, t)}{\sqrt{\balpha_t}})+\sqrt{1-\balpha_{t-1}}\boldeps_{\boldsymbol{\theta}}(\boldx_t, t)
\end{equation}

For conditional diffusion models, the most common practice is classifier-free guidance~\citep{ho2021classifier}, where the noise predictor is equipped and trained with an additional input condition $\boldc$, i.e. $\boldeps_{\boldsymbol{\theta}}(\boldx_t, t, \boldc)$. Here $\boldc$ can be a text embedding for Stable Diffusion~\citep{rombach2022high} or both text and image embeddings for Stable-unCLIP~\citep{ramesh2022hierarchical}. In recent years, training and sampling of diffusion models are conducted in the latent representation space~\citep{rombach2022high} instead of pixel space via a pair of pretrained encoder and decoder. Thus we use $\boldz\in\mcZ$ for latent codes and $\boldx\in\mcX$ for raw images.

%% file: section/3_2_method.tex
\subsection{\OURS}
\label{sec:method}

Given the advanced generative capability of diffusion models,
we seek to leverage them in risky sample generation. 
Although there are preliminary attempts to employ diffusion models to generate risky samples~\citep{chen2023advdiffuser,dai2024advdiff}, their methods lack attention to the conformity between generated samples and the desired category labels. 
When fixing a specific category $y$ and attempting to generate risky examples, we expect to generate samples that deceive a target model $f$ but still conform to the category $y$ from the view of humans. 
Previous methods adopt conditional diffusion models $\boldeps_{\boldtheta}(\boldz_t,t,y)$ with the desired category label $y$ (category index) as the input condition of the noise predictor, which is the only implicit supervision of conformity. 
The neglect of such conformity in the design of previous algorithms makes them fail to keep the ground-truth category label unchanged when achieving a high degree of risk for generated samples.

To mitigate this issue and further increase the degree of risk of generated samples, we propose \textbf{\OURS}. We adopt Stable-unCLIP $\boldeps_{\boldtheta}(\boldz_t, t, \boldc, \boldy_{text})$~\citep{ramesh2022hierarchical} that takes not only the desired category text embedding $\boldy_{text}$ but also a proper image embedding $\boldc$ as input conditions of the diffusion noise predictor, serving as implicit supervisions of conformity, and we add an explicit conformity constraint to keep the ground-truth label unchanged. We also design the mechanisms of embedding screening and risky gradient guidance to generate samples with higher risks. The following subsections are detailed descriptions of our algorithm, with a full procedure depicted in \Cref{fig:pipeline} and Algorithm \ref{alg:ours}.

\paragraph{Embedding Sampling and Screening}

Given a desired category $y$, for $\boldeps_{\boldtheta}(\boldz_t, t, \boldc, \boldy_{text})$, in each step of the diffusion sampling process, CLIP~\citep{radford2021learning} text embedding of the category name $\boldy_{text}$ is the input condition of the noise predictor, serving as an implicit constraint of conformity. 
For the other input condition, i.e. CLIP image embedding $\boldc$, we expect it to correspond to an image of the category $y$ so that it serves as another implicit constraint of conformity. 
We implement this via sampling from an estimated Gaussian distribution of the category. For the target model $f$ of a specific image classification task, its validation data $\{(\boldx_i,y_i)\}_{i=1}^n$ is often available. Thus we leverage these data to calculate the mean $\hat{\boldmu}_y$ and dimension-wise variance $\hat{\boldsigma}_y^2$ of their CLIP embeddings, and use them as the mean and diagonal elements of the covariance matrix for the Gaussian distribution. 
Since we want to generate risky samples, we design a simple embedding screening procedure to filter for the candidate embeddings that are more likely to generate samples on which $f$ makes mistakes. 
We fit an MLP as the error predictor $g_{\boldvarphi}$ using CLIP image embeddings of validation data and $e_i=\mathbb{I}(f(\boldx_i)\neq y)$, i.e. whether $f$ makes a mistake on $\boldx_i$. We do not require this error predictor to be very precise since it only serves as screening. 
Then we repeatedly sample $\boldc\sim \mathcal{N}(\hat{\boldmu}_y, {\rm diag}(\hat{\boldsigma}_y^2))$ until we have collected enough embeddings that are predicted by $g_{\boldvarphi}$ as $1$.

\begin{algorithm}[t]
\caption{\OURS} \label{alg:ours}
\begin{algorithmic}[1]
    \STATE \textbf{Input:} Tareget model $f$, unCLIP $\boldeps_{\boldtheta}$ along with its image embedder $h$ and image decoder $\mcD$, validation dataset $\{(\boldx_i,y_i)\}_{i=1}^n$, category name embedding $\boldy_{text}$ and category label $y$, gradient scale $s$, conformity coefficient $\lambda$.
    \STATE Calculate image embeddings: $\boldc_i=h(\boldx_i),\ \forall i\in [n]$.
    \STATE Calculate model errors: $e_i=\mathbb{I}(f(\boldx_i)\neq y_i),\ \forall i\in [n]$. 
    \STATE Fit an error predictor $g_{\boldvarphi}$ with $\{(\boldc_i,e_i)\}_{i=1}^n$.
    \STATE Calculate $\hat{\boldmu}_y$ as the mean of $\{\boldc_i|y_i=y,i\in [n]\}$.
    \STATE Calculate $\hat{\boldsigma}_y^2$ whose $d$-th dimension is the variance of $\{c_{i,d}|y_i=y,i\in [n]\}$.
    \STATE Repeat sampling $\boldc\sim\mathcal{N}(\hat{\boldmu}_y, {\rm diag}(\hat{\boldsigma}_y^2))$ till $g_{\boldvarphi}(\boldc)=1$. 
    \STATE Sample $\boldz_T\sim\mathcal{N}(\boldsymbol{0},\boldI)$
    \FOR{$t=T$ to $1$}
    \STATE $\hat{\boldx}=\mcD(\hat{\boldz}_0)=\mcD(\frac{\boldz_t-\sqrt{1-\balpha_t}\boldeps_{\boldtheta}(\boldz_t, t, \boldc, \boldy_{text})}{\sqrt{\balpha_t}})$.
    \STATE $\hat{S}=\ell(f(\hat{\boldx}),y)+\lambda h(\hat{\boldx})\cdot \boldy_{text}$.
    \STATE $\hat{\boldeps}=\boldeps_{\boldtheta}(\boldz_t, t, \boldc, \boldy_{text})-s\sqrt{1-\balpha_t}\frac{\nabla_{\boldz_t}\hat{S}}{||\nabla_{\boldz_t}\hat{S}||}$.
    \STATE $\boldz_{t-1}=\sqrt{\balpha_{t-1}}(\frac{\boldz_t-\sqrt{1-\balpha_t}\hat{\boldeps}}{\sqrt{\balpha_t}})+\sqrt{1-\balpha_{t-1}}\hat{\boldeps}$.
    \ENDFOR
    \STATE \textbf{return:} $\mcD(\boldz_0)$.
\end{algorithmic}
\end{algorithm}

\paragraph{Risky Gradient Guidance}

To fool the classifier $f$ so that it does not predict the generated sample as $y$, inspired by classifier-guided conditional generation~\citep{dhariwal2021diffusion}, we incorporate the gradient of $f$ into the sampling process of Stable-unCLIP. 
Specifically, for step $t$, we adopt the rough estimate $\hat{\boldz}_0=\frac{\boldz_t-\sqrt{1-\balpha_t}\boldeps_{\boldtheta}(\boldz_t, t, \boldc, \boldy_{text})}{\sqrt{\balpha_t}}$ and decode it as $\hat{\boldx}$. Then we calculate the risk score $\hat{S}=\ell(f(\hat{\boldx}), y)$ as the loss between the output of $f$ and $y$. For DDIM, we incorporate the gradient of the risk score into noise prediction following~\citet{dhariwal2021diffusion}:
\begin{equation}
\label{eq:adv}
    \hat{\boldeps}=\boldeps_{\boldtheta}(\boldz_t, t, \boldc, \boldy_{text})-s\sqrt{1-\balpha_t}\frac{\nabla_{\boldz_t}\hat{S}}{||\nabla_{\boldz_t}\hat{S}||}   
\end{equation}
where $s$ is the scale of gradient guidance. This could guide the sampling process of the diffusion model towards the direction where $\hat{S}$ is larger, i.e. $f(\hat{\boldx})$ is less likely to be $y$.

\paragraph{Conformity Gradient Guidance}

To further guarantee the conformity between the generated sample and the conditioned category $y$, we introduce an explicit constraint into the sampling process of diffusion. 
In addition to $\ell(f(\hat{\boldx}), y)$, we add the inner product of CLIP image embedding of $\hat{\boldx}$ and the CLIP text embedding of the category name, i.e. $h(\hat{\boldx})\cdot \boldy_{text}$, to the guidance score. 
This enhances their conformity in the representation space of CLIP. Thus the score could be rewritten as:
\begin{equation}
\label{eq:score}
    \hat{S}=\ell(f(\hat{\boldx}),y)+\lambda h(\hat{\boldx})\cdot \boldy_{text}
\end{equation}
where $\lambda$ is conformity coefficient, $h$ is CLIP image encoder, $\boldy_{text}$ is CLIP text embedding of the category name. Note that since CLIP image and text encoders are both already available during training and sampling of Stable-unCLIP, our adoption of them does not require extra resources.

%% file: section/4_exp.tex
\section{Experiments}
\label{sec:exp}

\subsection{Experimental Settings}

Here we provide key details of experimental settings. More details can be found in \Cref{appendix:detail}. 

\paragraph{Datasets} We conduct our experiments on four image classification datasets: CIFAR-100~\citep{krizhevsky2009learning}, ImageNet~\citep{deng2009imagenet}, PACS~\citep{li2017deeper}, and NICO++~\citep{zhang2023nico++}. The former two are well-known datasets in standard image classification, while the latter two are widely used image classification datasets in domain generalization,
each with multiple domains or environments, so that we could evaluate both in-distribution (ID) generalization and out-of-distribution (OOD) generalization performance when utilizing generated risky samples to augment training data.

\paragraph{Target models} 
For each dataset, we generate risky examples for four target models. 
For CIFAR-100, we employ DenseNet-121~\citep{huang2017densely}, ResNet-50~\citep{he2016deep}, WideResNet~\citep{zagoruyko2016wide}, and GoogleNet~\citep{szegedy2015going}, trained using code from a popular github repository\footnote{https://github.com/weiaicunzai/pytorch-cifar100}. 
For ImageNet, we directly use DenseNet-121, ResNet-50, EfficientNet-B2~\citep{tan2019efficientnet}, and ViT-B/16~\citep{dosovitskiy2020image} trained on ImageNet provided by torchvision. 
For PACS and NICO++, we train models of the same backbones as those of ImageNet using code from the domain generalization benchmark DomainBed~\citep{gulrajanisearch2021}. 
More specifically, for PACS, we train on the domain of art and leave domains of photo, cartoon, and sketch as OOD domains to evaluate OOD generalization performance after augmenting training data with generated risky samples. For NICO++, we train on the domains of autumn, rock, dim, and grass, and leave outdoor and water as OOD domains.

\paragraph{Baselines} In our experiments of risky sample generation, we mainly compare with AdvDiffuser~\citep{chen2023advdiffuser} and AdvDiff~\citep{dai2024advdiff}, which also adopt diffusion models to generate risky samples. 
For the experiments of generalization enhancement via augmenting training data with generated risky samples, we also compare with the classic data augmentation technique Mixup~\citep{zhang2017mixup} and straightforward utilization of Stable-unCLIP~\citep{ramesh2022hierarchical} as a data augmenter.

\begin{table*}[htbp]
\centering
\caption{Error rate (\%) of generated samples on CIFAR-100, ImageNet, PACS, and NICO++ with respect to various target models. Higher is better. We can see that our method \OURS~consistently achieves the highest error rate in all settings. }
\label{tab:asr}
\resizebox{0.9\textwidth}{!}{%
\begin{tabular}{@{}c|cccc|cccc@{}}
\toprule
Dataset      & \multicolumn{4}{c|}{CIFAR-100}                                  & \multicolumn{4}{c}{ImageNet}                                    \\ \midrule
Target Model & DenseNet-121  & ResNet-50     & WideResNet      & GoogleNet     & DenseNet-121  & ResNet-50     & EfficientNet-B2 & ViT-B/16      \\ \midrule
AdvDiffuser  & 83.1          & 77.3          & 77.5            & 81.8          & 51.2          & 65.1          & 53.5            & 68.9          \\
AdvDiff      & 44.5          & 44.1          & 47.2            & 46.1          & 15.0          & 12.5          & 23.2            & 11.4          \\
Ours         & \textbf{83.5} & \textbf{78.5} & \textbf{85.5}   & \textbf{83.2} & \textbf{68.2} & \textbf{71.5} & \textbf{70.8}   & \textbf{71.4} \\ \midrule
Dataset      & \multicolumn{4}{c|}{PACS}                                       & \multicolumn{4}{c}{NICO++}                                      \\ \midrule
Target Model & DenseNet-121  & ResNet-50     & EfficientNet-B2 & ViT-B/16      & DenseNet-121  & ResNet-50     & EfficientNet-B2 & ViT-B/16      \\ \midrule
AdvDiffuser  & 15.1          & 31.2          & 15.6            & 23.4          & 35.0          & 38.6          & 28.0            & 35.2          \\
AdvDiff      & 17.3          & 16.4          & 14.5            & 33.0          & 19.2          & 16.5          & 13.9            & 12.8          \\
Ours         & \textbf{19.3} & \textbf{56.9} & \textbf{19.8}   & \textbf{34.6} & \textbf{39.6} & \textbf{40.2} & \textbf{41.3}   & \textbf{39.2} \\ \bottomrule
\end{tabular}%
}
\end{table*}
\begin{figure*}[h]
  \centering
  \includegraphics[width=\linewidth]{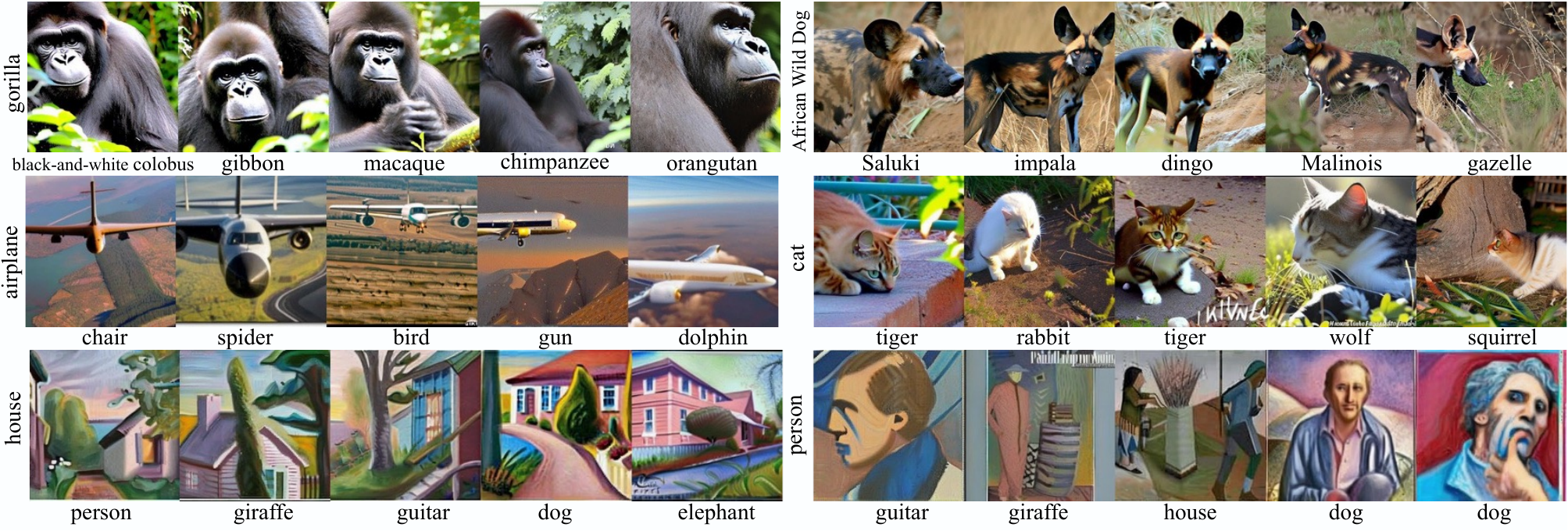}
  \caption{Risky samples generated by~\OURS. Three rows correspond to ImageNet, NICO++, and PACS, each including two categories. The left caption implies ground truth category. The lower caption imply prediction of the target model (ResNet-50).}
  \label{fig:showcase}
\end{figure*}

\paragraph{Other details} For PACS, we set the gradient scale $s=20$. For the other three datasets, we set $s=10$. We set the conformity coefficient $\lambda=1e-4$.
For detailed hyperparameter analyses, please refer to \Cref{appendix:ablation}. 
For hyperparameters of baselines, we directly follow their own settings.  
As for the number of generated images, for ImageNet we generate 20 images for each category, resulting in a total of 20,000 samples for each method. For CIFAR-100, PACS, and NICO++, we generate 120 images for each category, resulting in 12,000, 840, and 7,200 samples for each method.

\subsection{Experimental Results}

We conduct experiments from multiple perspectives to comprehensively illustrate the superiority of our method.

\subsubsection{Degree of risk} 

We compare the degree of risk, measured by the error rate of generated samples, of our method with baselines on four target models for four datasets. 
From \Cref{tab:asr}, we can see that our method significantly increases the degree of risk of generated samples over previous methods across various target models and various datasets. In ImageNet, PACS, and NICO++, our method increases the error rate by a large margin. 
This provides strong evidence of our method's superior capability in generating risky samples that could deceive the given model. 
We provide examples generated by our method in \Cref{fig:showcase}. 
We can see that many of our generated risky samples are semantically interpretable. E.g., 
for the category ``gorilla", the lighting and shadow in the face of the gorilla in the first image bear resemblance to the black and white furs of ``black-and-white colobus". 
For the category ``cat", the color and stripes of the cat resemble those of a tiger, which could be the reason for misclassification.
Thus our method could also help explore and interpret possible mistakes that a model might make in prediction. 
For more examples, please refer to \Cref{appendix:examples}.

\subsubsection{Generation quality}

Another important aspect of risky sample generation is the quality of generated images. 
In previous literature~\citep{chen2023advdiffuser,dai2024advdiff}, they directly employ FID scores as the evaluation metric that is also commonly adopted in image generation tasks. 
However, when the distribution of generated images is closer to the original data distribution, a lower FID could also be achieved, while we want the distribution of generated images to be closer to that of the data on which the target model predicts falsely. 
Therefore, we calculate the FID between generated images and the original error samples. 
We adopt the implementation of clean-FID
~\citep{parmar2022aliased} to calculate FID scores. 
From \Cref{tab:fid}, we can see that our method consistently achieves a lower FID score than baselines in every setting. This indicates that our method is capable of generating images with higher quality.

\subsubsection{Transferability}

\begin{table}[h]
\centering
\caption{Transferability of generated risky samples. The error rate (\%) is measured on other models using samples generated for ResNet-50. Higher is better.}
\label{tab:transfer}
\resizebox{0.4\textwidth}{!}{%
\begin{tabular}{@{}c|ccc@{}}
\toprule
Dataset      & \multicolumn{3}{c}{CIFAR-100}                   \\ \midrule
Target Model & DenseNet-121  & WideResNet      & GoogleNet     \\ \midrule
AdvDiffuser  & 44.7          & 43.9            & 44.8          \\
AdvDiff      & 41.1          & 41.6            & 42.4          \\
\OURS~(Ours)         & \textbf{62.5} & \textbf{62.2}   & \textbf{61.0} \\ \midrule
Dataset      & \multicolumn{3}{c}{ImageNet}                    \\ \midrule
Target Model & DenseNet-121  & EfficientNet-B2 & ViT-B/16      \\ \midrule
AdvDiffuser  & 32.9          & 30.6            & 29.7          \\
AdvDiff      & 8.1           & 7.4             & 6.1           \\
\OURS~(Ours)         & \textbf{42.6} & \textbf{40.2}   & \textbf{39.0} \\ \bottomrule
\end{tabular}%
}
\end{table}

\begin{table*}[htbp]
\centering
\caption{FID scores between generated samples and original error samples on CIFAR-100, ImageNet, PACS, and NICO++. Lower is better. Our method consistently achieves the lowest FID in all settings, indicating the highest image generation quality. }
\label{tab:fid}
\resizebox{0.9\textwidth}{!}{%
\begin{tabular}{@{}c|cccc|cccc@{}}
\toprule
Dataset      & \multicolumn{4}{c|}{CIFAR-100}                                  & \multicolumn{4}{c}{ImageNet}                                    \\ \midrule
Target Model & DenseNet-121  & ResNet-50     & WideResNet      & GoogleNet     & DenseNet-121  & ResNet-50     & EfficientNet-B2 & ViT-B/16      \\ \midrule
AdvDiffuser  & 45.2          & 45.0          & 45.2            & 45.1          & 26.4          & 27.1          & 25.4            & 23.1          \\
AdvDiff      & 45.3          & 45.0          & 44.1            & 45.2          & 13.5          & 15.7          & 14.9            & 13.8          \\
\OURS~(Ours)         & \textbf{33.2} & \textbf{32.9} & \textbf{34.2}   & \textbf{33.4} & \textbf{10.2} & \textbf{12.4} & \textbf{10.4}   & \textbf{11.5} \\ \midrule
Dataset      & \multicolumn{4}{c|}{PACS}                                       & \multicolumn{4}{c}{NICO++}                                      \\ \midrule
Target Model & DenseNet-121  & ResNet-50     & EfficientNet-B2 & ViT-B/16      & DenseNet-121  & ResNet-50     & EfficientNet-B2 & ViT-B/16      \\ \midrule
AdvDiffuser  & 67.4          & 72.3          & 78.4            & 70.5          & 25.6          & 25.1          & 27.2            & 26.5          \\
AdvDiff      & 67.8          & 72.6          & 78.9            & 75.4          & 26.1          & 25.9          & 26.1            & 26.4          \\
\OURS~(Ours)         & \textbf{47.9} & \textbf{67.0} & \textbf{54.6}   & \textbf{46.1} & \textbf{17.3} & \textbf{17.5} & \textbf{23.2}   & \textbf{18.4} \\ \bottomrule
\end{tabular}%
}
\end{table*}

Although our generation process is conducted with respect to a specific target model, we conduct experiments to verify that our generated risky samples could transfer well to other models. 
In \Cref{tab:transfer}, we can see that samples generated by \OURS~for ResNet-50 still achieve a high error rate on other models and outperform those of other methods on CIFAR-100 and ImageNet. 
More results can be found in~\Cref{appendix:transfer}.

\begin{figure}[h]
  \centering
  \includegraphics[width=\linewidth]{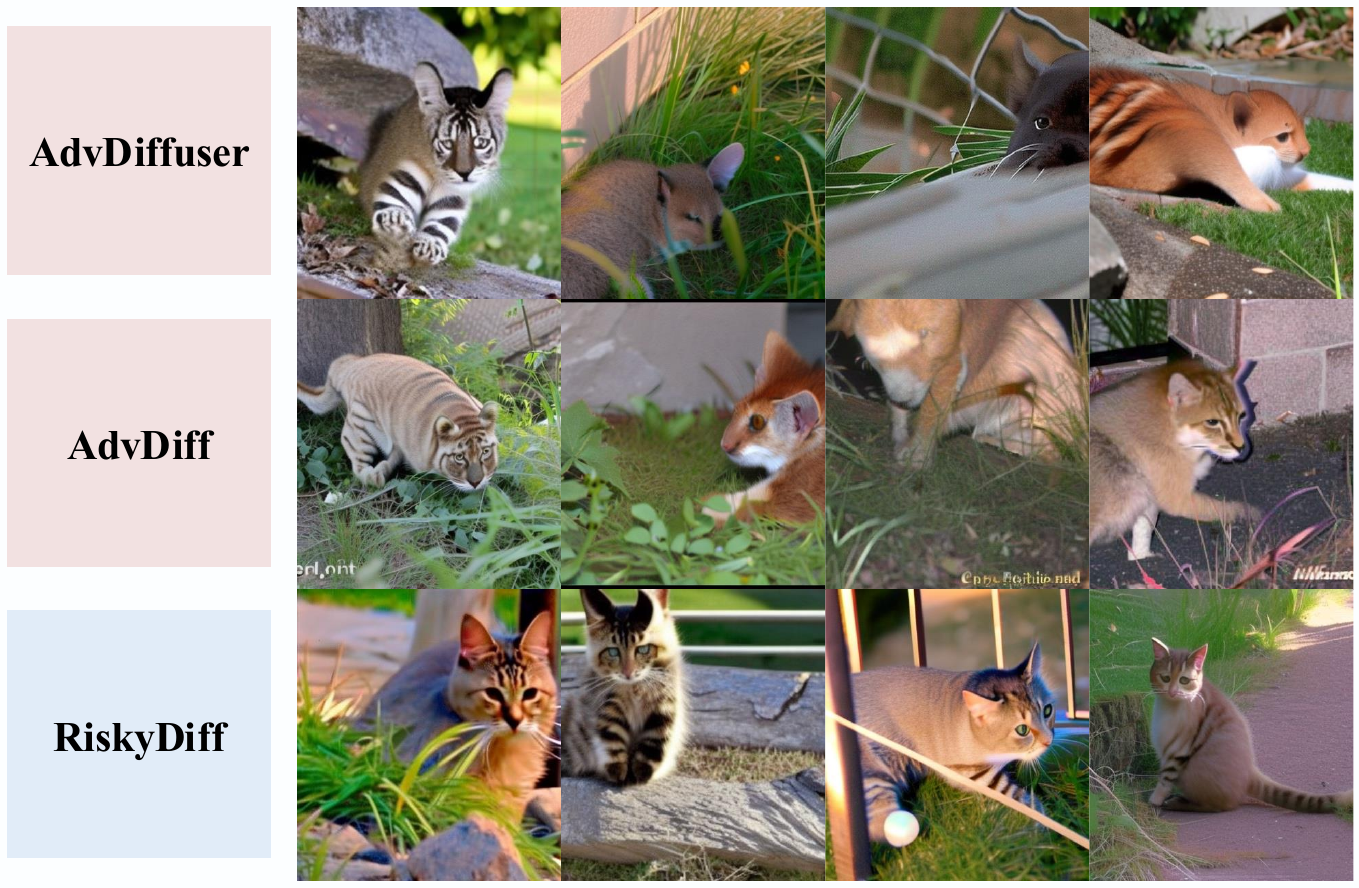}
  \caption{Images of cats generated by AdvDiffuser, AdvDiff, and our method with ResNet-50 as the target model on NICO++. It shows that AdvDiffuser and AdvDiff fail to preserve the true characteristics of cats while our method can.}
  \label{fig:conform}
\end{figure}

\subsubsection{Generation conformity} 
A fundamental aspect of risky sample generation is the conformity between the generated samples and the expected category labels. 
In a standard generation process via an advanced diffusion model, the conformity between the generated samples and their expected category labels could mostly be guaranteed. 
However, when attempting to generate risky samples, the image generation process is guided towards difficult or rare cases, where the diffusion model might not be guaranteed to keep the expected category label unchanged. 
Thus it is necessary to conduct extra analyses on the generation conformity. 
From \Cref{fig:conform}, for the category ``cat", we can see that some images generated by AdvDiffuser and AdvDiff are not really cats already. For example, the first image of AdvDiffuser looks more like a tiger instead of a cat. 
Meanwhile, the conformity of our generated samples could also be validated from experiments of generalization enhancement in the next paragraph.

\subsubsection{Generalization enhancement}

A critical but less investigated aspect is how the generated or perturbed samples could be used to further help the target models correct their potential mistakes and enhance their generalization ability on natural examples. 
Therefore, we retrain the models on CIFAR-100, PACS, and NICO++ by adding the generated data to the original training data as augmentation and evaluate their test performance. Each retraining process is repeated three times with different random seeds. 
For CIFAR-100, we measure the accuracy on the test set. For domain generalization datasets PACS and NICO++, we measure the ID accuracy on a holdout validation set of the training domains and OOD accuracy on all the other domains. 
In addition to AdvDiffuser and AdvDiff, we also compare with the classic data augmentation technique Mixup~\citep{zhang2017mixup} and direct utilization of images generated by Stable-unCLIP~\citep{ramesh2022hierarchical}.

\begin{table}[h]
\centering
\caption{Accuracy of various methods on CIFAR-100. For unCLIP, AdvDiffuser, AdvDiff, and \OURS, we add their generated images to training data and retrain. }
\label{tab:acc-cifar}
\resizebox{0.48\textwidth}{!}{%
\begin{tabular}{@{}c|cccc@{}}
\toprule
Model       & DenseNet-121     & ResNet-50        & WideResNet       & GoogleNet        \\ \midrule
Original    & 79.1             & 78.4             & 78.9             & 77.0             \\ \midrule
Mixup       & 79.5$_{\pm 0.2}$ & 78.9$_{\pm 0.1}$ & 79.9$_{\pm 0.1}$ & 76.9$_{\pm 0.1}$ \\
unCLIP      & 79.5$_{\pm 0.3}$ & 79.4$_{\pm 0.4}$ & 79.3$_{\pm 0.2}$ & 77.2$_{\pm 0.2}$ \\
AdvDiffuser & 79.0$_{\pm 0.2}$ & 78.8$_{\pm 0.1}$ & 79.1$_{\pm 0.3}$ & 76.3$_{\pm 0.3}$ \\
AdvDiff     & 79.0$_{\pm 0.2}$ & 78.4$_{\pm 0.0}$ & 79.2$_{\pm 0.3}$ & 76.5$_{\pm 0.1}$ \\
\OURS~(Ours)        & \textbf{79.8}$_{\pm 0.1}$ & \textbf{79.7}$_{\pm 0.0}$ & \textbf{80.4}$_{\pm 0.2}$       & \textbf{77.7}$_{\pm 0.1}$ \\ \bottomrule
\end{tabular}%
}
\end{table}

From \Cref{tab:acc-cifar}, we can see that our generated images successfully improve the test accuracy of different models on CIFAR-100 compared with the original models, and our method outperforms baselines of risky sample generation and other data augmentation techniques in terms of the performance after retraining. 
We also notice that images generated by AdvDiffuser and AdvDiff could decrease the performance of the original model under some settings. This serves as additional evidence that the nonconformity between images generated by these methods and their expected category labels is more severe than ours, thus bringing stronger label noise and hurting the model performance.

\begin{table*}[htbp]
\centering
\caption{ID and OOD accuracy of various methods on PACS. For unCLIP, AdvDiffuser, AdvDiff, and \OURS, it implies that we add their generated images to the training data and retrain. }
\label{tab:acc-pacs}
\resizebox{0.85\textwidth}{!}{%
\begin{tabular}{@{}c|cc|cc|cc|cc@{}}
\toprule
Model       & \multicolumn{2}{c|}{DenseNet-121}   & \multicolumn{2}{c|}{ResNet-50}      & \multicolumn{2}{c|}{EfficientNet-B2} & \multicolumn{2}{c}{ViT-B/16}        \\ \midrule
Method      & ID Acc.          & OOD Acc.         & ID Acc.          & OOD Acc.         & ID Acc.           & OOD Acc.         & ID Acc.          & OOD Acc.         \\ \midrule
Original    & 94.7            & 73.6            & 93.7            & 77.7            & 96.0             & 76.3            & 94.4            & 72.3            \\ \midrule
Mixup       & 94.6$_{\pm 0.3}$ & 71.1$_{\pm 1.7}$ & 95.2$_{\pm 0.4}$ & 72.4$_{\pm 1.4}$ & 96.1$_{\pm 0.4}$  & 73.8$_{\pm 3.5}$ & 96.5$_{\pm 0.3}$ & 74.2$_{\pm 0.7}$ \\
unCLIP      & 94.8$_{\pm 0.3}$ & 73.6$_{\pm 1.8}$ & 94.3$_{\pm 0.6}$ & 74.4$_{\pm 1.1}$ & 95.6$_{\pm 0.0}$  & 78.3$_{\pm 0.7}$ & 95.8$_{\pm 0.3}$ & 73.6$_{\pm 2.8}$ \\
AdvDiffuser & 94.4$_{\pm 0.2}$ & 75.3$_{\pm 2.9}$ & 94.5$_{\pm 0.6}$ & 74.2$_{\pm 2.2}$ & 95.7$_{\pm 0.3}$  & 78.5$_{\pm 1.1}$ & 95.1$_{\pm 0.4}$ & 73.9$_{\pm 3.8}$ \\
AdvDiff     & 93.8$_{\pm 0.4}$ & 70.8$_{\pm 0.4}$ & 94.5$_{\pm 0.2}$ & 74.4$_{\pm 0.7}$ & 95.8$_{\pm 0.2}$  & 78.3$_{\pm 1.0}$ & 95.7$_{\pm 0.1}$ & 75.0$_{\pm 0.4}$ \\
\OURS~(Ours)        & \textbf{96.0}$_{\pm 0.1}$ & \textbf{77.8}$_{\pm 0.8}$ & \textbf{95.8}$_{\pm 0.1}$ & \textbf{81.4}$_{\pm 1.0}$ & \textbf{96.5}$_{\pm 0.1}$  & \textbf{82.8}$_{\pm 0.5}$ & \textbf{96.7}$_{\pm 0.1}$ & \textbf{79.1}$_{\pm 0.1}$ \\ \bottomrule
\end{tabular}%
}
\end{table*}
\begin{table*}[htbp]
\centering
\caption{ID and OOD accuracy of various methods on NICO++. For unCLIP, AdvDiffuser, AdvDiff, and \OURS, it implies that we add their generated images to the training data and retrain. }
\label{tab:acc-nico}
\resizebox{0.85\textwidth}{!}{%
\begin{tabular}{@{}c|cc|cc|cccc@{}}
\toprule
Model       & \multicolumn{2}{c|}{DenseNet-121}   & \multicolumn{2}{c|}{ResNet-50}      & \multicolumn{2}{c|}{EfficientNet-B2}                     & \multicolumn{2}{c}{ViT-B/16}        \\ \midrule
Method      & ID Acc.          & OOD Acc.         & ID Acc.          & OOD Acc.         & ID Acc.          & \multicolumn{1}{c|}{OOD Acc.}         & ID Acc.          & OOD Acc.         \\ \midrule
Original    & 82.5             & 69.0             & 82.9             & 69.6             & 85.8             & \multicolumn{1}{c|}{73.1}             & 86.0             & 74.3             \\ \midrule
Mixup       & 84.1$_{\pm 0.4}$ & 71.5$_{\pm 0.4}$ & 84.2$_{\pm 0.1}$ & 70.7$_{\pm 0.2}$ & 86.8$_{\pm 0.1}$ & \multicolumn{1}{c|}{75.0$_{\pm 0.2}$} & 87.0$_{\pm 0.1}$ & 74.5$_{\pm 0.7}$ \\
unCLIP      & 82.9$_{\pm 0.2}$ & 69.6$_{\pm 0.2}$ & 83.3$_{\pm 0.1}$ & 70.0$_{\pm 0.2}$ & 86.2$_{\pm 0.1}$ & \multicolumn{1}{c|}{73.4$_{\pm 0.1}$} & 84.8$_{\pm 2.0}$ & 73.7$_{\pm 0.5}$ \\
AdvDiffuser & 81.4$_{\pm 2.1}$ & 69.1$_{\pm 0.1}$ & 81.5$_{\pm 2.0}$ & 69.4$_{\pm 0.3}$ & 81.9$_{\pm 0.2}$ & \multicolumn{1}{c|}{73.5$_{\pm 0.5}$} & 82.3$_{\pm 0.3}$ & 74.9$_{\pm 0.7}$ \\
AdvDiff     & 77.9$_{\pm 0.2}$ & 68.4$_{\pm 0.6}$ & 80.1$_{\pm 2.1}$ & 69.7$_{\pm 0.4}$ & 84.7$_{\pm 2.0}$ & \multicolumn{1}{c|}{73.3$_{\pm 0.3}$} & 82.1$_{\pm 0.1}$ & 74.6$_{\pm 0.3}$ \\
\OURS~(Ours)        & \textbf{85.9}$_{\pm 0.2}$ & \textbf{72.7}$_{\pm 0.0}$ & \textbf{85.1}$_{\pm 0.0}$ & \textbf{71.5}$_{\pm 0.2}$ & \textbf{88.2}$_{\pm 0.0}$ & \textbf{76.2}$_{\pm 0.1}$                      & \textbf{87.8}$_{\pm 0.1}$ & \textbf{75.4}$_{\pm 0.4}$ \\ \bottomrule
\end{tabular}%
}
\end{table*}

\begin{table}[htb]
\centering
\caption{Error rate and FID scores for ResNet-50 on NICO++ with varying size of validation data. ``Val. Size" represents the used fraction of the original validation data.}
\label{tab:size}
\resizebox{0.45\textwidth}{!}{%
\begin{tabular}{@{}c|cccccccc@{}}
\toprule
Val. Size & 0.1  & 0.2  & 0.3  & 0.5  & 0.7  & 0.8  & 0.9  & 1    \\ \midrule
Error rate      & 39.4 & 39.6 & 39.6 & 39.8 & 40.2 & 40.6 & 41.0 & 41.2 \\
FID       & 17.9 & 17.9 & 17.7 & 17.7 & 17.7 & 17.5 & 17.5 & 17.5 \\ \bottomrule
\end{tabular}%
}
\end{table}

\begin{figure}[H]
  \centering
  \includegraphics[width=0.9\linewidth]{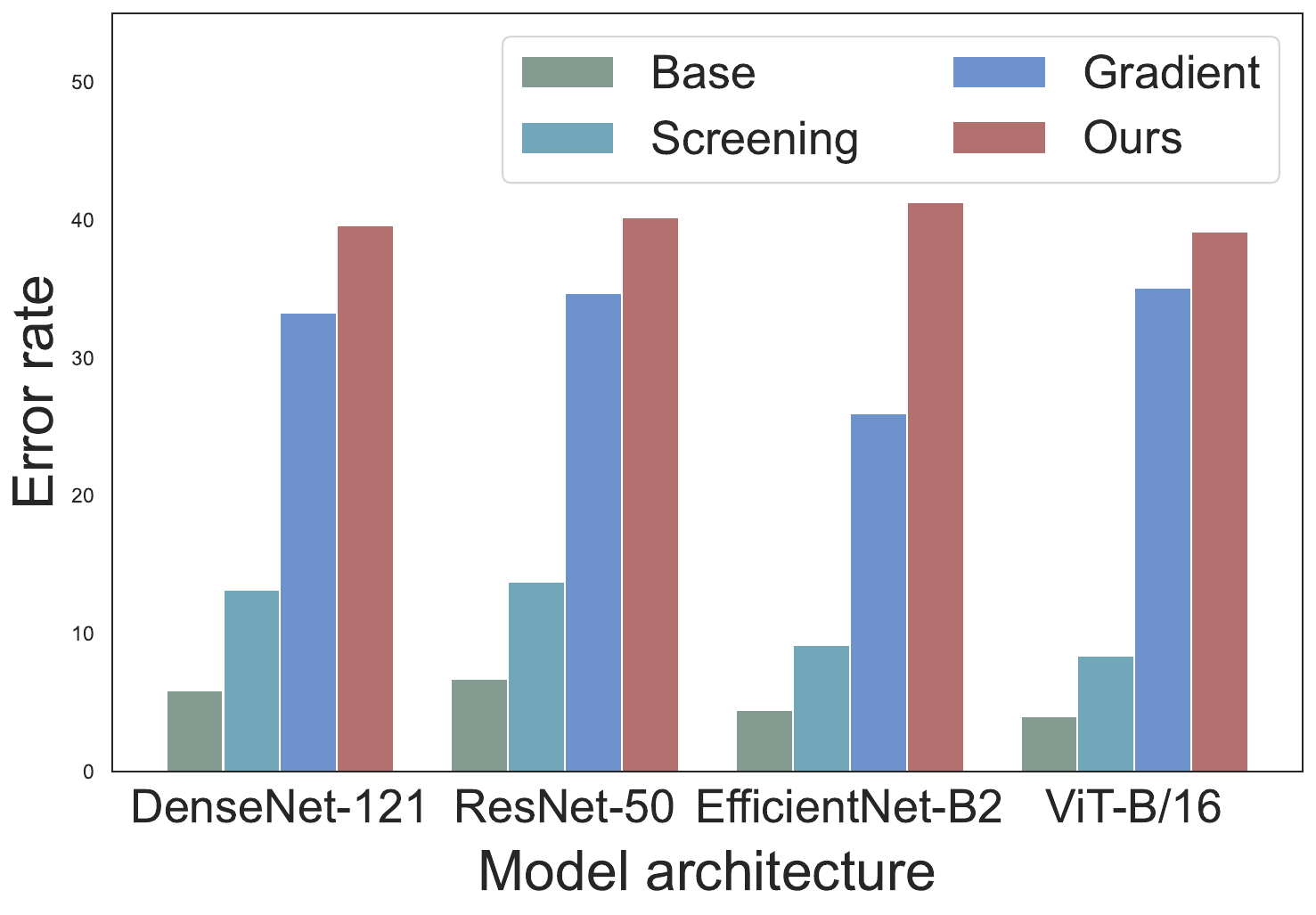}
  \caption{Ablation study of embedding screening and gradient guidance on NICO++. 
``Base" is direct generating images using Stable-unCLIP. ``Screening" is embedding screening. ``Gradient'' is gradient guidance.  }
  \label{fig:ablation}
\end{figure}

From \Cref{tab:acc-pacs} and \ref{tab:acc-nico}, we can see that our generated images increase both ID and OOD generalization performance by a large margin compared with the original models, which demonstrates the enhancement of both ID and OOD generalization abilities brought by our generated images, and our method consistently outperforms other methods on various model backbones. Similar to CIFAR-100, we also observe a decrease of performance for AdvDiffuser and AdvDiff compared with the original models across many settings. This further indicates the possible label noise brought by these methods due to the nonconformity between generated examples and category labels, and signifies the higher conformity of our generated examples.

\subsubsection{Ablation Study}

To verify the usefulness of different components of our method, we conduct an ablation study of embedding screening and gradient guidance on NICO++ with different target models. 
\Cref{fig:ablation} shows that each of the individual components could increase the error rate, and the combination outperforms each individual one. 
Meanwhile, to understand the reliability of validation data, we conduct experiments by reducing the size of validation data gradually to $1/10$ of the original size. From~\Cref{tab:size}, we can see that when decreasing validation data size, both error rate and FID change by only a small margin. 
This indicates that our method is effective even with a small size of validation data.

%% file: section/5_con.tex
\section{Conclusion}
\label{sec:con}

In this paper, 
to overcome the challenge of keeping conformity between generated risky samples and expected category labels, we introduce \OURS~by integrating both image and text embeddings as implicit constraints of category conformity, and propose an explicit conformity constraint via a conformity score. 
We also design the mechanisms of embedding screening and risky gradient guidance. 
Experiments demonstrate that \OURS~significantly outperforms baselines in terms of error rate, generation quality, and category conformity, and illustrate that the generated risky samples of high conformity could be utilized to enhance the generalization capability of the target models.

%% file: section/X_suppl.tex
\section{Appendix}

\subsection{Experimental Details}
\label{appendix:detail}

In this subsection, we provide details that are not mentioned in our main paper due to the space limit. 

\paragraph{Datasets} We list detailed information of our employed datasets here. 
\begin{itemize}
    \item CIFAR-100~\citep{krizhevsky2009learning}: A widely-adopted image classification dataset with a relatively low resolution of $32\times 32$. It has 100 categories. There are 50,000 images for training and 10,000 images for testing. 
    \item ImageNet~\citep{deng2009imagenet}: A well-known and challenging image classification dataset whose most commonly used version has 1,000 categories. It has 1,281,167 training images and 50,000 validation images. 
    \item PACS~\citep{li2017deeper}: A popular image classification dataset in domain generalization. It has 7 categories and 4 domains of different styles (photo, art painting, cartoon, sketch). When training on some domains and testing on the other domains, it simulates a setting of distribution shift. 
    \item NICO++~\citep{zhang2023nico++}: An image classification dataset of 60 categories in domain generalization with 10 domains. We use the 6 publicly available domains in our experiments: autumn, rock, dim, grass, outdoor, water, thus it generates out-of-distribution scenarios via various background contexts. 
\end{itemize}

\paragraph{Baselines and Implementation} 
The main baselines we have compared with are AdvDiff~\citep{dai2024advdiff} and  AdvDiffuser~\citep{chen2023advdiffuser} that also employ diffusion models to generate risky samples. 
Since AdvDiff has released the source code, we directly use it. Since AdvDiffuser has not released the source code, we implement it by ourselves. 
For each method, we generate and save the examples, and run the same test code to measure the error rate for a fair comparison.

\subsection{Hyperparameter Analyses}
\label{appendix:ablation}

\begin{table}[htbp]
\centering
\caption{Hyperparameter analyses of the gradient scale $s$ and the conformity coefficient $\lambda$. Experiments are conducted on NICO++ for ResNet-50. }
\label{tab:hp}
\resizebox{0.4\textwidth}{!}{%
\begin{tabular}{@{}ccc|ccc@{}}
\toprule
\multicolumn{3}{c|}{$\lambda=0.0001$}  & \multicolumn{3}{c}{$s=10$}                   \\ \midrule
\multicolumn{1}{c|}{$s$} & Error rate  & FID  & \multicolumn{1}{c|}{$\lambda$} & Error rate  & FID  \\ \midrule
\multicolumn{1}{c|}{0.1} & 6.9  & 17.8 & \multicolumn{1}{c|}{0}         & 41.3 & 19.7 \\
\multicolumn{1}{c|}{1}   & 8.6  & 17.8 & \multicolumn{1}{c|}{0.00001}   & 42.0 & 18.6 \\
\multicolumn{1}{c|}{5}   & 21.4 & 17.7 & \multicolumn{1}{c|}{0.00005}   & 40.9 & 18.0 \\
\multicolumn{1}{c|}{10}  & 40.2 & 17.5 & \multicolumn{1}{c|}{0.0001}    & 40.2 & 17.5 \\
\multicolumn{1}{c|}{20}  & 58.8 & 18.4 & \multicolumn{1}{c|}{0.001}     & 36.3 & 17.3 \\
\multicolumn{1}{c|}{30}  & 65.7 & 19.6 & \multicolumn{1}{c|}{0.01}      & 27.5 & 17.3 \\
\multicolumn{1}{c|}{50}  & 73.8 & 20.3 & \multicolumn{1}{c|}{0.1}       & 15.9 & 17.2 \\ \bottomrule
\end{tabular}%
}
\end{table}

\begin{figure}[h]
	\centering
	\begin{subfigure}{0.2\linewidth}
            \includegraphics[width=\linewidth]{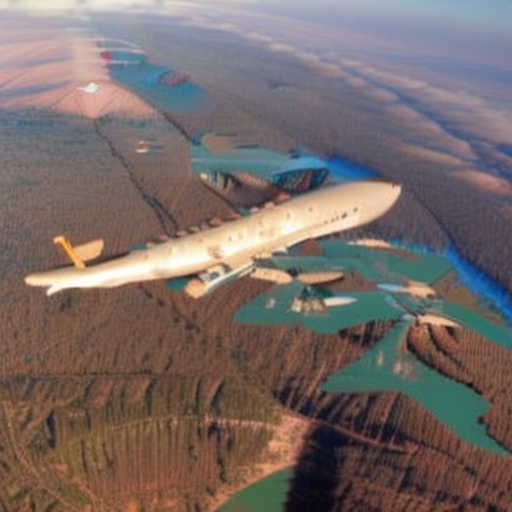}
	    \caption{$s=10$}
            \label{fig:gs1}
	\end{subfigure} 
	\begin{subfigure}{0.2\linewidth}
            \includegraphics[width=\linewidth]{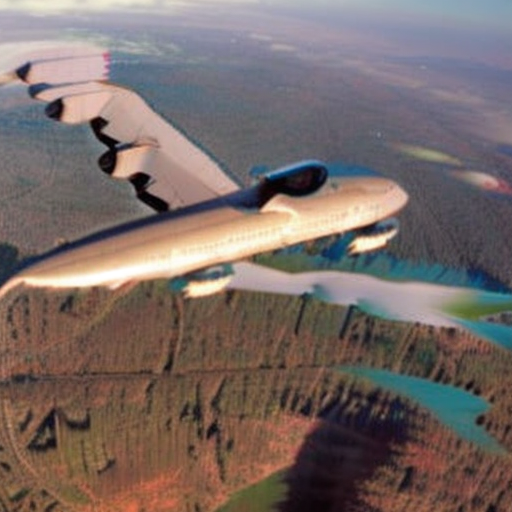}
	    \caption{$s=20$}
            \label{fig:gs2}
	\end{subfigure} 
	\begin{subfigure}{0.2\linewidth}
            \includegraphics[width=\linewidth]{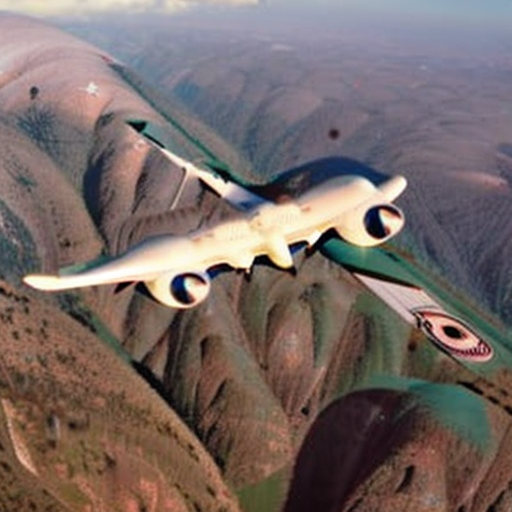}
	    \caption{$s=50$.}
            \label{fig:gs3}
	\end{subfigure}
	\caption{Examples of an airplane when fixing $\lambda=0.0001$ and increasing $s$. We can see that when $s>20$, it starts to look strange instead of maintaining the characteristics of an airplane. }
	\label{fig:gs}
\end{figure}
\begin{figure}[h]
	\centering
	\begin{subfigure}{0.2\linewidth}
            \includegraphics[width=\linewidth]{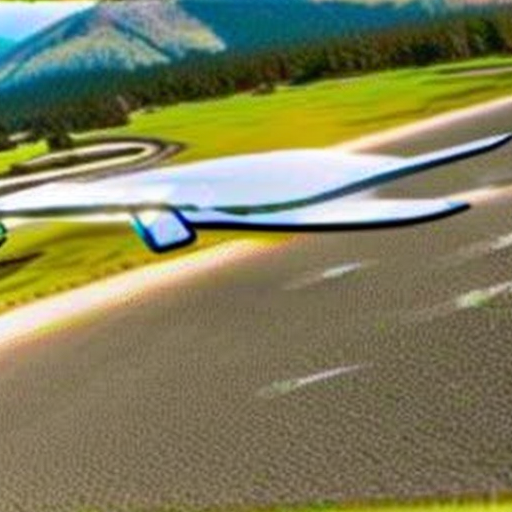}
	    \caption{$\lambda=0$}
            \label{fig:ms1}
	\end{subfigure} 
	\begin{subfigure}{0.2\linewidth}
            \includegraphics[width=\linewidth]{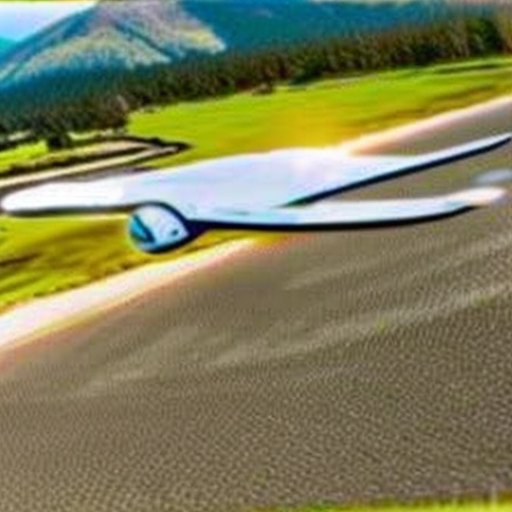}
	    \caption{$\lambda=0.00001$}
            \label{fig:ms2}
	\end{subfigure} 
	\begin{subfigure}{0.2\linewidth}
            \includegraphics[width=\linewidth]{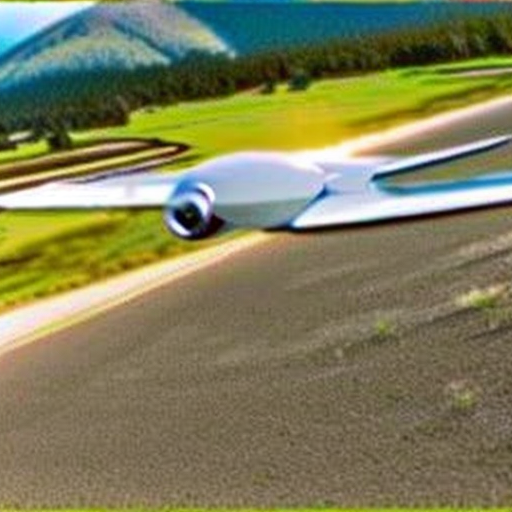}
	    \caption{$\lambda=0.0001$.}
            \label{fig:ms3}
	\end{subfigure}
	\caption{Examples of an airplane when fixing $s=10$ and increasing $\lambda$. We can see that when $\lambda=0$, the generated example does not look very much like an airplane. When $\lambda$ increases, the generated example exhibits more realistic features of an airplane. }
	\label{fig:ms}
\end{figure}

In this subsection, we conduct hyperparameter analyses of the gradient scale $s$ and the conformity coefficient $\lambda$ on NICO++ for ResNet-50. 
In \Cref{tab:hp}, when fixing $\lambda$, we can see that when $s\geq 10$, the error rate becomes competitive. However, when $s> 20$, we check and find that there are many more examples that fail to preserve conformity to the expected category. 
We show an example in \Cref{fig:gs}. 
Meanwhile, we find that $s=10$ achieves the lowest FID. Note that the FID scores adopted in this paper are calculated between generated samples and error examples. Thus when increasing the strength of risky gradient guidance, it is possible that the FID score might decrease since the distribution of generated samples might get closer to that of error samples. 
So we choose $s=10$ for datasets with a large number of categories like NICO++, CIFAR-100, and ImageNet. 
For PACS whose number of categories is only 7, we empirically increase $s$ to $20$ for riskier generated samples. 
When fixing $s$, we can see that when $\lambda\leq 0.0001$, conformity gradient guidance only decreases the error rate by a small range. 
We manually check and find that increasing $s$ truly helps in increasing the conformity between generated examples and their corresponding categories. 
We show an example in \Cref{fig:ms}. 
Thus we set $\lambda=0.0001$.

\subsection{More Examples}
\label{appendix:examples}

\begin{figure*}[htbp]
  \centering
  \includegraphics[width=\linewidth]{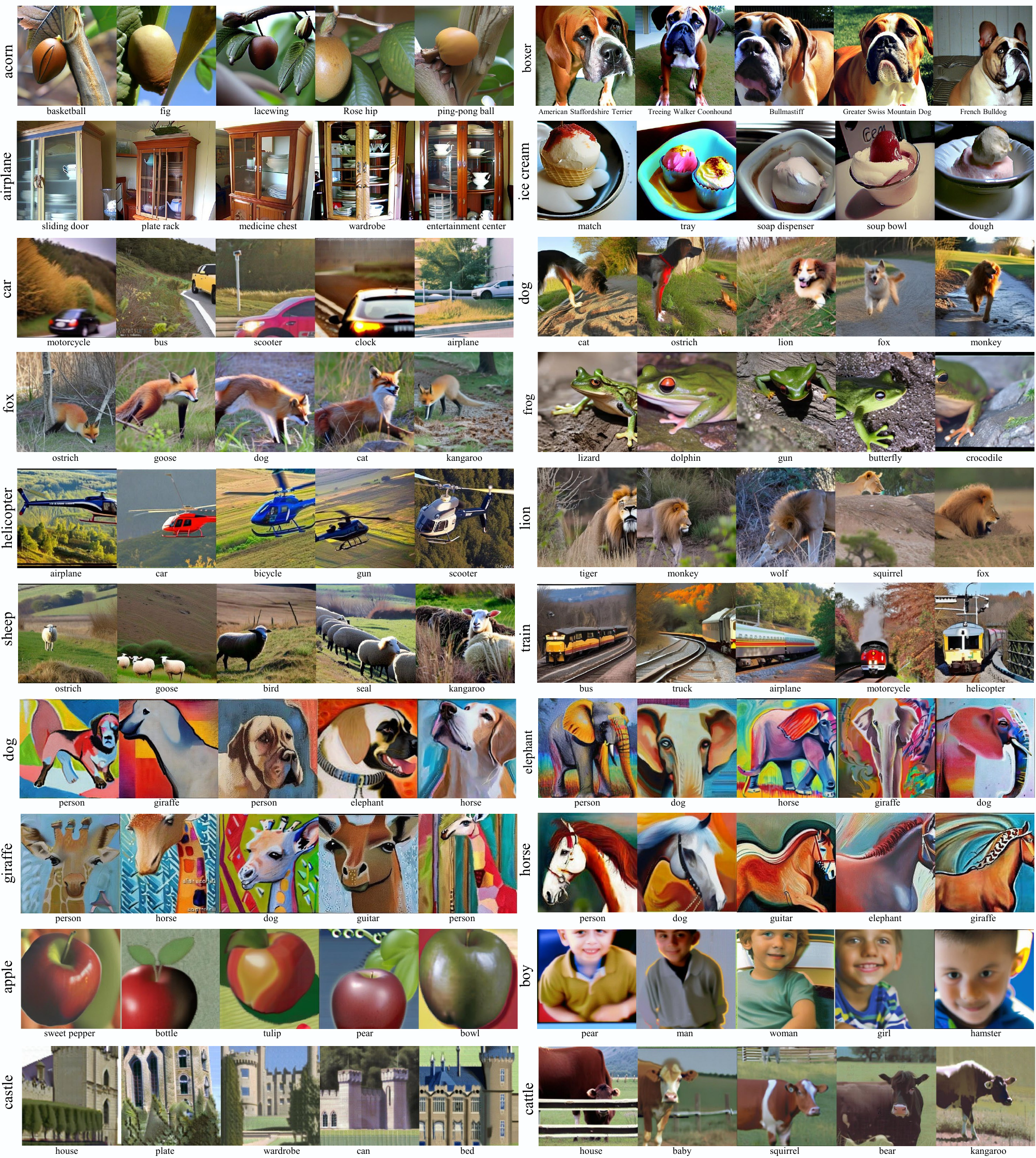}
  \caption{More risky samples generated by our method. First 2 rows correspond to ImageNet. Row 3$\sim$6 correspond to NICO++. Rows 7$\sim$8 correspond to PACS. Last 2 rows correspond to CIFAR-100. Each row includes two categories from the dataset, with the left caption implying the ground truth category, and the lower caption representing the prediction of the target model (ResNet-50).}
  \label{fig:showcase-appendix}
\end{figure*}

In this subsection, we provide more examples generated by our method. From \Cref{fig:showcase-appendix}, we can see that our method successfully generates risky samples that conform to their expected categories.

\subsection{Discussion on OOD Generalization}

It is worth noting that an important downstream application of risky sample generation is to augment training data to further improve the Out-of-Distribution (OOD) generalization ability of models~\citep{liu2021towards}, as revealed in~\Cref{tab:acc-pacs} and~\ref{tab:acc-nico}. Previously literature attempt to propose algorithms from various perspectives to improve the OOD generalization ability, including invariant learning~\citep{liu2021heterogeneous, liu2025environment, liu2024inductive, creager2021environment}, stable learning~\citep{kuang2020stable,shen2020stable,yu2023stable,yu2025sample}, domain generalization~\citep{cha2021swad,rame2022fishr,zhang2023gradient,zhang2023flatness}, etc.
Recently, a new paradigm suggests that instead of directly designing OOD algorithms, it could be better to first comprehensively evaluate the OOD generalization ability of models~\cite{wu2024bridging,yu2024survey, blanchet2024stability,yu2024rethinking,yu2025odp} and then collect data with guidance from findings of evaluation to improve model generalization in a data-centric style~\citep{liu2023need,he2024domain,zha2025data}, to which our work bears resemblance.

\clearpage

\clearpage

\subsection{Transferability}
\label{appendix:transfer}

\Cref{tab:appendix-transfer} shows results of transferability on the four datasets. We can see that our method consistently achieves a high error rate and outperforms other methods in almost all settings.

\begin{table*}[h]
\centering
\caption{Transferability of generated risky samples. The error rate (\%) is measured on other models using samples generated for ResNet-50. Higher is better.}
\label{tab:appendix-transfer}
\resizebox{0.9\textwidth}{!}{%
\begin{tabular}{@{}c|ccc|ccc@{}}
\toprule
Dataset      & \multicolumn{3}{c|}{CIFAR-100}                  & \multicolumn{3}{c}{ImageNet}                    \\ \midrule
Target Model & DenseNet-121  & WideResNet      & GoogleNet     & DenseNet-121  & EfficientNet-B2 & ViT-B/16      \\ \midrule
AdvDiffuser  & 44.7          & 43.9            & 44.8          & 32.9          & 30.6            & 29.7          \\
AdvDiff      & 41.1          & 41.6            & 42.4          & 8.1           & 7.4             & 6.1           \\
\OURS~(Ours)         & \textbf{62.5} & \textbf{62.2}   & \textbf{61.0} & \textbf{42.6} & \textbf{40.2}   & \textbf{39.0} \\ \midrule
Dataset      & \multicolumn{3}{c|}{PACS}                       & \multicolumn{3}{c}{NICO++}                      \\ \midrule
Target Model & DenseNet-121  & EfficientNet-B2 & ViT-B/16      & DenseNet-121  & EfficientNet-B2 & ViT-B/16      \\ \midrule
AdvDiffuser  & 16.1          & 12.5            & 12.1          & 10.2          & 8.1             & \textbf{7.8}  \\
AdvDiff      & 15.7          & 13.2            & 12.9          & 9.1           & 7.4             & 7.7           \\
\OURS~(Ours)         & \textbf{34.8} & \textbf{46.8}   & \textbf{30.7} & \textbf{12.7} & \textbf{8.4}    & 6.4           \\ \bottomrule
\end{tabular}%
}
\end{table*}

\subsection{Time Comparison}
\label{appendix:time}

We conduct running time comparisons for different methods on NICO++. From \Cref{tab:time}, we can see that the time cost of our method is slightly higher than previous methods. Considering its effectiveness in generating risky samples with conformity, such an increase of time cost is worthwhile. 
Time comparison experiments are all run on the same Linux server, where CPU is AMD EPYC 7402 24-Core Processor, and all the 8 GPUs are NVIDIA GeForce RTX 3090. 

\begin{table}[htbp]
\centering
\caption{Running time comparison for different methods. The time represents the generation time per image, and is measured in second. Experiments are conducted on NICO++. }
\label{tab:time}
\resizebox{0.65\textwidth}{!}{%
\begin{tabular}{@{}c|cccc@{}}
\toprule
Method           & DenseNet-121 & ResNet-50 & EfficientNet-B2 & ViT-B/16 \\ \midrule
AdvDiffuser      & 7.79         & 7.35      & 8.15            & 7.89     \\
AdvDiff          & 8.87         & 8.55      & 8.70            & 8.77     \\
\OURS~(Ours) & 9.25         & 8.99      & 9.29            & 9.02     \\ \bottomrule
\end{tabular}%
}
\end{table}

\subsection{Code}

Code is available at \url{https://github.com/h-yu16/RiskyDiff}.